\documentclass[journal, letterpaper]{IEEEtran}
\usepackage[T1]{fontenc}

\newcommand{\maximize}{\mathop{\rm maximize}}
\newcommand{\minimize}{\mathop{\rm minimize}}
\newcommand{\subjecto}{\mathop{\rm subject\;to}}

\newtheorem{theorem}{Theorem}

\newtheorem{assumption}{Assumption}

\newcommand{\argmax}{\mathop{\rm argmax}}

\usepackage{cite}
\ifCLASSINFOpdf
  \usepackage[pdftex]{graphicx}
  \usepackage[unicode,hidelinks]{hyperref}
\else
  \usepackage[dvips]{graphicx}
  \DeclareGraphicsExtensions{.eps}
\fi
\usepackage[cmex10]{amsmath}
\usepackage{amssymb}
\usepackage{mathtools}
\usepackage[ruled]{algorithm}
\usepackage{algpseudocode}
\usepackage{mathrsfs}
\usepackage[caption=false,font=footnotesize]{subfig}
\usepackage{fixltx2e}
\usepackage{makecell}
\usepackage{multirow}
\usepackage{hhline}
\usepackage{enumitem}
\usepackage{microtype}
\usepackage{color}
\usepackage[cmintegrals]{newtxmath}
\ifCLASSOPTIONcaptionsoff
  \usepackage[nomarkers]{endfloat}
 \let\MYoriglatexcaption\caption
 \renewcommand{\caption}[2][\relax]{\MYoriglatexcaption[#2]{#2}}
\fi
\usepackage{kotex}
\usepackage{setspace}
\usepackage{etoolbox}
\AtBeginEnvironment{algorithmic}{\onehalfspacing}

\newcommand{\calN}{\ensuremath{{\mathcal{N}}}}
\newcommand{\calM}{\ensuremath{{\mathcal{M}}}}
\newcommand{\calR}{\ensuremath{{\mathcal{R}}}}

\newcommand{\calL}{\ensuremath{{\mathcal{L}}}}
\newcommand{\calS}{\ensuremath{{\mathcal{S}}}}
\newcommand{\calA}{\ensuremath{{\mathcal{A}}}}
\newcommand{\calP}{\ensuremath{{\mathcal{P}}}}

\newcommand{\bW}{\ensuremath{{\mathbf W}}}

\newcommand{\bw}{\ensuremath{{\mathbf w}}}
\newcommand{\bx}{\ensuremath{{\mathbf x}}}

\newcommand{\bq}{\ensuremath{{\mathbf q}}}
\newcommand{\bQ}{\ensuremath{{\mathbf Q}}}

\newcommand{\btheta}{\ensuremath{{\boldsymbol \uptheta}}}

\newcommand{\blambda}{\ensuremath{{\boldsymbol \uplambda}}}

\makeatletter
\newcommand{\algmargin}{\the\ALG@thistlm}
\makeatother
\newlength{\whilewidth}
\algdef{SE}[parWHILE]{parWhile}{EndparWhile}[1]
  {\parbox[t]{\dimexpr\linewidth-\algmargin}{%
     \hangindent\whilewidth\strut\algorithmicwhile\ #1\ \algorithmicdo\strut}}{\algorithmicend\ \algorithmicwhile}%
\algnewcommand{\parState}[1]{\State%
  \parbox[t]{\dimexpr\linewidth-\algmargin}{\strut #1\strut}}

\usepackage{suffix}
\usepackage{mathtools}

\begin{document}
\title{Collaborative Policy Learning for Dynamic Scheduling Tasks in Cloud-Edge-Terminal IoT Networks Using Federated Reinforcement Learning}

\author{Do-Yup~Kim,~\IEEEmembership{Member,~IEEE,}~Da-Eun~Lee,~Ji-Wan~Kim,~and~Hyun-Suk~Lee
\thanks{D.-Y. Kim is with the Department of Information and Communication AI Engineering, Kyungnam University, Changwon-si, Gyeongsangnam-do 51767, South Korea (e-mail: doyup09@kyungnam.ac.kr).}
\thanks{D.-E. Lee, J.-W. Kim, and H.-S. Lee are with the School of Intelligent Mechatronics Engineering, Sejong University, Seoul, South Korea (e-mail: kjs990516@naver.com, jiwan1228@naver.com, hyunsuk@sejong.ac.kr).}
}

\maketitle

\begin{abstract}
In this paper, we examine cloud-edge-terminal IoT networks, where edges undertake a range of typical dynamic scheduling tasks.
In these IoT networks, a central policy for each task can be constructed at a cloud server.
The central policy can be then used by the edges conducting the task, thereby mitigating the need for them to learn their own policy from scratch.
Furthermore, this central policy can be collaboratively learned at the cloud server by aggregating local experiences from the edges, thanks to the hierarchical architecture of the IoT networks.
To this end, we propose a novel collaborative policy learning framework for dynamic scheduling tasks using federated reinforcement learning.
For effective learning, our framework adaptively selects the tasks for collaborative learning in each round, taking into account the need for fairness among tasks.
In addition, as a key enabler of the framework, we propose an edge-agnostic policy structure that enables the aggregation of local policies from different edges.
We then provide the convergence analysis of the framework.
Through simulations, we demonstrate that our proposed framework significantly outperforms the approaches without collaborative policy learning. Notably, it accelerates the learning speed of the policies and allows newly arrived edges to adapt to their tasks more easily.\looseness=-1
\end{abstract}


\bstctlcite{IEEEexample:BSTcontrol}

\section{Introduction}

\begin{figure}[!t]
	\centering
	\includegraphics[width=.9\linewidth]{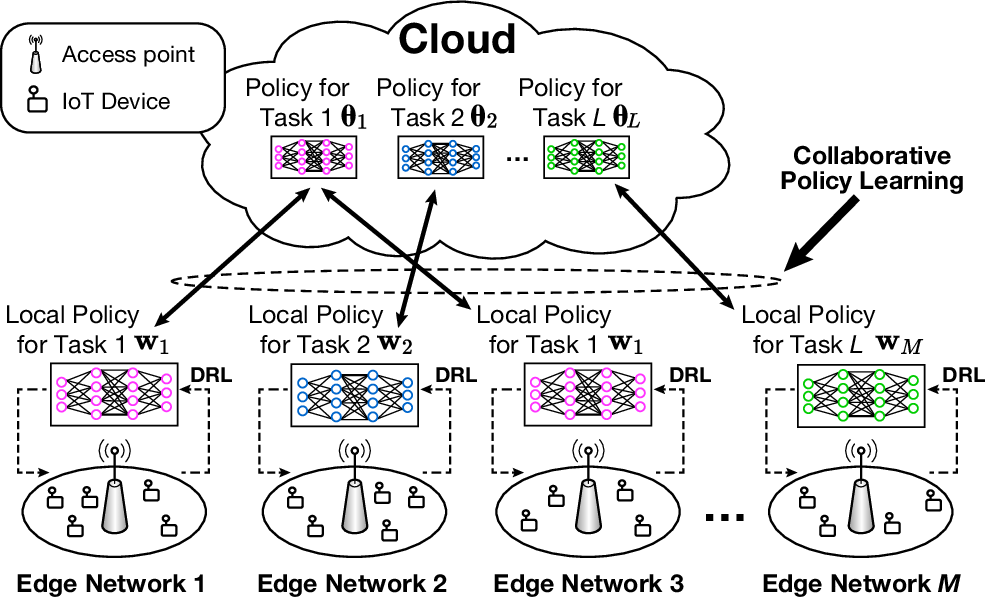}
	\caption{Architecture of cloud-edge-terminal IoT networks with collaborative policy learning for dynamic scheduling tasks.\label{fig:arch}}
	\vspace{-1em}
\end{figure}

With the recent explosive development of internet-of-things (IoT) applications, a hierarchical architecture for IoT networks has been widely studied to ensure agility, flexibility, and scalability \cite{pan2017future,qiu2020networking,wang2021eihdp}.
In this hierarchical architecture, IoT networks can be decomposed into edges and a cloud-edge network, as illustrated in Fig. \ref{fig:arch}.
Each edge forms its own network, called an edge network, comprising an access point (AP) and IoT terminal devices, while the cloud-edge network consists of a cloud server and edge networks.
Such hierarchical IoT networks are typically referred to as cloud-edge-terminal IoT networks, as they consist of a cloud server, edges, and IoT terminal devices.\looseness=-1

In this hierarchical architecture, edges in IoT networks carry out numerous \textit{tasks}, such as inference, prediction, planning, and scheduling, to support various IoT applications and services.
In particular, a variety of \textit{dynamic scheduling tasks} have been widely considered as major tasks in IoT networks.
Dynamic scheduling tasks typically involve a problem, where an item is chosen from multiple items to achieve a goal, which has been widely considered in various applications, from recommendation \cite{shani2005mdp, huang2021deep, lu2016partially} to resource scheduling \cite{wei2018user, zhu2021joint, ye2018deep, xu2017deep, han2008resource, kim2022low, lee2022radio} to queueing \cite{ferra2003applying, stidham1993survey, chang2000line, park2022joint}.
In IoT networks, different edge functionalities, such as radio resource management \cite{lee2019resource, malik2018radio, he2018green, shi2020deep}, data gathering \cite{peng2020learning, zhang2020hierarchical}, and wireless power transfer \cite{lee2019contextual, xiong2020uav, lee2020adaptive}, correspond to this problem, in which an edge selects an IoT terminal device from multiple IoT terminal devices for the corresponding functionalities.
It is worth emphasizing that in typical IoT networks, multiple edges share common dynamic scheduling tasks since these functionalities are generally used in IoT networks.
For example, most edge nodes should conduct radio resource management tasks to serve IoT terminal devices. Additionally, in most sensor applications, each edge node carries out data aggregation scheduling tasks, which schedule IoT terminal devices to effectively aggregate data from each IoT terminal device.\looseness=-1

To efficiently address dynamic scheduling tasks in IoT networks, deep learning, especially deep reinforcement learning (DRL), has been widely applied \cite{he2018green, shi2020deep, lee2019resource, peng2020learning, zhang2020hierarchical, lee2019contextual, xiong2020uav, lee2020adaptive}.
DRL is one of the representative methods for solving complex stochastic problems, thanks to the large representational capability of deep learning.
Specifically, in DRL-based approaches, an agent directly learns a policy represented by a deep neural network (DNN) model to address its task using data or experiences obtained from interactions with environments.
Consequently, these approaches allow each edge to find policies for its tasks without the formulation and optimization of complex scheduling task problems based on hand-crafted mathematical models, as in traditional approaches.\looseness=-1

In cloud-edge-terminal IoT networks, a cloud server can play the role of coordinator to manage policies for tasks, thanks to the hierarchical architecture.
Therefore, it may be possible that a central policy for each dynamic scheduling task can be constructed at a cloud server.
Then, newly arrived edges can avoid performance deterioration due to an initial learning phase by using the central policy instead of learning its own policy.
Besides, with the coordination of the cloud server, the edges that conduct the task can cooperate in learning the central policy so as to learn the policy more efficiently.
One intuitive way for such cooperation is to directly collect \textit{data} (i.e., experiences) from the edges to the cloud server.
The cloud server then learns a policy to solve the problem using the collected data and redistributes the policy to the edges.
However, this approach is impractical since directly uploading the data from edges to the cloud server causes privacy and security issues \cite{yang2019federated}.
Moreover, it incurs unaffordable communication costs due to the transmission of an enormous amount of data from all the edges to the cloud server \cite{lim2020federated}.\looseness=-1

As a viable solution to address these issues, federated learning has been widely studied \cite{yang2019federated, lim2020federated}.
In federated learning, a cloud server and local learners sharing an identical task can cooperate to efficiently train a central DNN model to address the task.
Specifically, in each round, each local learner trains its local DNN model using its local training data and uploads its trained local DNN model to the cloud server instead of its local data.
The cloud server can then improve the central model by aggregating the received local models and redistributing it to the local learners.
This enables the central model to be trained in a distributed manner while avoiding privacy issues.
By applying this procedure to DRL, federated reinforcement learning (FRL) has also been studied \cite{wang2020federated, yu2020deep}.
We refer the reader to comprehensive surveys of federated learning in \cite{lim2020federated} for more details.\looseness=-1

The hierarchical architecture of cloud-edge-terminal IoT networks is suitable for applying federated learning to collaboratively learn a policy for each task.
Since multiple edges share an identical task, each edge can act as a local learner for the policy of the task, and the cloud server can aggregate the local policies of the edges.
Thus, federated learning in cloud-edge-terminal IoT networks has been studied for tasks such as mobile keyboard prediction, cyberattack detection, and energy demand prediction \cite{lim2020federated}.
%
However, there is no work yet on FRL frameworks that enables edges to collaboratively address their \textit{dynamic scheduling tasks}, even though a variety of works for dynamic scheduling tasks have been studied based on DRL.
This is because conventional DRL-based approaches that have been studied so far are inapplicable to FRL.
Specifically, they are developed to learn a policy focused on achieving its goal only in a target edge.
Consequently, the policy focuses on addressing the characteristics of the target edge, such as the number of IoT terminal devices and the statistics of system uncertainties, rather than generalizing them for application to all edges.
Furthermore, the conventional DRL-based approaches make the corresponding policies for different target edges have different structures, even if their tasks are identical.
As a result, it is difficult to aggregate the policies learned from different edges via FRL due to their dependency on \textit{edge-specific} characteristics.
Therefore, to enable edges to collaboratively learn policies for dynamic scheduling tasks, a novel policy structure is needed which can be used for any different edges while avoiding such edge-specific characteristics.\looseness=-1

Even if FRL can be applied for collaborative policy learning for dynamic scheduling tasks in cloud-edge-terminal IoT networks, it is difficult to simply use it because of the scarcity of cloud resources, such as computing power, memory, and network bandwidth \cite{weingartner2015cloud}.
The larger the number of edges participating in FRL for collaborative policy learning, the greater the usage of cloud resources, making it harder to aggregate local policies for all tasks. 
Moreover, while the larger number of edges participating in FRL generally improves the efficiency of FRL due to the increased amount of experiences \cite{xia2020multi, lee2021adaptive}, some edges may not be available to participate in FRL in each round.
Hence, to maximize the effectiveness of collaborative policy learning, the tasks whose local policies are to be aggregated in each round should be carefully selected to effectively utilize limited cloud resources while considering the following factors: the number of available edges for each task in the round and the number of edges that have participated in FRLg of each task so far.
However, there is no such work on collaborative policy learning frameworks for tasks in cloud-edge-terminal IoT networks yet.


In this paper, we study collaborative policy learning for dynamic scheduling tasks in cloud-edge-terminal IoT networks.
In these IoT networks, edges share a variety of dynamic scheduling tasks, as illustrated in Fig. \ref{fig:arch}.
Specifically, each edge conducts its own dynamic scheduling tasks and learns the policies for the corresponding tasks using DRL.
Meanwhile, a cloud server trains the central policy for each task by aggregating the local policies that are learned at different edges via FRL.
To this end, we propose a collaborative policy learning framework for dynamic scheduling tasks in cloud-edge-terminal IoT networks.
In this framework, a cloud server manages the policy for each dynamic scheduling task, which is commonly conducted across multiple edges, and learns it by aggregating the local policies for the task from the edges.
This collaborative learning process accelerates the learning speed of the policy for each task in the cloud-edge-terminal IoT networks.
Additionally, when new edges arrive in the network, they can easily adapt to conducting their tasks by utilizing the central policies for the tasks.\looseness=-1

The contributions of this paper are summarized as follows:
\begin{itemize}
\item We propose a novel collaborative policy learning framework for dynamic scheduling tasks in IoT networks using FRL. It learns the central policy for each task, which is edge-agnostic, by effectively utilizing limited cloud resources and considering the uncertainties in the availability of participating edges for FRL. We provide a convergence analysis of the proposed framework.\looseness=-1

\item In the proposed framework, we develop a task selection algorithm that adaptively selects the tasks for which local policies are to be federated. This enhances the effectiveness of learning the central policies. Specifically, it aims to maximize the total average number of edges that participate in FRL, while considering fairness among tasks. As a result, this approach facilitates the effective learning of central policies for all tasks.\looseness=-1

\item As an enabler of the proposed framework, we propose an edge-agnostic policy structure for a given dynamic scheduling task, which is applicable to collaborative policy learning. It possesses the capability to generalize the edge-specific characteristics of the policy for the task. Consequently, local policies based on this edge-agnostic policy structure can be well aggregated by FRL.\looseness=-1

\item Through extensive experiments, we demonstrate that the proposed framework enables cloud-edge-terminal IoT networks to learn the policies for dynamic scheduling tasks in a distributed manner. Thanks to this, it achieves significant performance improvement compared with the approaches that do not utilize collaborative policy learning. In addition, our framework provides adaptability for newly arrived edges and accelerates the learning speed of the policy.\looseness=-1

\end{itemize}

The rest of this paper is organized as follows. Section \ref{sec:model_problem} presents the system model and problem formulation for dynamic scheduling tasks. Section \ref{sec:challenges} discusses some key challenges in the context of collaborative policy learning, and Section \ref{sec:FL_framework} presents a collaborative policy learning framework designed to address these challenges. In Section \ref{sec:experimental_results}, we present experimental results to validate the effectiveness of the proposed framework. Finally, Section \ref{sec:conclusion} provides the conclusion of the paper.

\section{Cloud-Edge-Terminal IoT Networks With Multiple Dynamic Scheduling Tasks}
\label{sec:model_problem}

\subsection{System Model of Cloud-Edge-Terminal IoT Networks}
\label{sec:model_iot_net}
We consider a cloud-edge-terminal IoT network\footnote{For brevity, we will henceforth refer to ``cloud-edge-terminal IoT network'' simply as ``IoT network'' throughout the paper.} that consists of a cloud server and multiple edges, each with multiple IoT devices, as illustrated in Fig. \ref{fig:arch}.
We denote the set of edges by $\calN=\{1,2,\ldots,N\}$, where $N$ is the number of edges, and edge $n\in\calN$ is composed of one access point (AP) and $M_n$ IoT devices.
The set of IoT devices in edge $n$ is denoted by $\calM_n=\{1,2,\ldots,M_n\}$.
Each edge carries out one of several general dynamic scheduling tasks,\footnote{For brevity, we will interchangeably use ``dynamic scheduling task'' and ``task'' throughout the paper if there is no confusion.} such as radio resource management \cite{malik2018radio, he2018green, shi2020deep, lee2019resource}, data gathering \cite{peng2020learning, zhang2020hierarchical}, and wireless power transfer \cite{lee2019contextual, xiong2020uav, lee2020adaptive}.\footnote{It is worth noting that this system model offers a straightforward extension to scenarios where an edge carries out multiple tasks. This can be accomplished by conceptualizing the edge as a collection of distinct virtual edges, with each one representing an individual task.}
We postulate the existence of $L$ distinct types of tasks, and we denote the set of these tasks by $\calL=\{1,2,\ldots,L\}$.
Here, each element $l \in \calL$ signifies a unique individual task.
We proceed to denote the task of edge $n$ as $l(n)\in\calL$.
Additionally, we define the set of edges involved in task $l$ as $\calN(l) = \{n:l(n)=l\}$.
Lastly, we consider the maximum network bandwidth $W$, memory resource $O$, and computing resource $C$ of the IoT network for performing FRL in the cloud server.

\subsection{Dynamic Scheduling Tasks in Edges}
\label{sec:task_model}

We now describe various types of dynamic scheduling tasks, as provided in the previous subsection, using the following common procedure.
Each edge selects an IoT device and makes decisions relevant to scheduling (e.g., transmission power in wireless network scheduling and the number of jobs to be serviced in job scheduling) to achieve the goal of the respective task.
Additionally, each edge considers each IoT device's conditions relevant to scheduling (e.g., the current queue length in queue scheduling and the channel conditions in wireless network scheduling) for effective scheduling.

From this procedure, we formulate a \textit{generic} dynamic scheduling problem structure for edges that can represent various types of dynamic scheduling tasks.
To this end, we first provide a system model for each edge $n$ performing its corresponding task $l(n)$.
Each edge is assumed to performs its task over a discrete time horizon $t\in\{1,2,\ldots\}$.
It is worth noting that the time horizon is defined individually for each edge to describer its task, and it does not imply a global time horizon that spans across multiple edges.
Continuing, we define the state information vector of IoT device $m\in\calM_n$ in time slot $t$ by $s_{n,m}^t=(s_{n,m,1}^t, \ldots, s_{n,m,K(l(n))}^t)$, where $s_{n,m,k}^t$ is the $k$th state information of IoT device $m$ in time slot $t$, and $K(l)$ is the number of types of state information for task $l$.
Then, we can define a state of edge $n$ in time slot $t$ as
\begin{equation}
	\label{eqn:task_state_definition}
	s_n^t=(s_{n,1}^t,\ldots,s_{n,M_n}^t)\in\calS_n,
\end{equation}
where $\calS_n$ is the state space.
We also define an action of edge $n$ in time slot $t$ as
\begin{equation}
	\label{eqn:task_action_definition}
	a_n^t=(m_n^t,g_1^t,\ldots,g_{G(l(n))}^t)\in\calA_n,
\end{equation}
where $m_n^t\in\calM_n$ is the IoT device scheduled by edge $n$ in time slot $t$, and $\{g_1^t, \ldots ,g_{G(l(n))}^t\}$ represent the decision set relevant to scheduling, where $G(l)$ is the number of relevant decisions for task $l$.
Next, we let $u_l(s,a)$ be the reward function for task $l$, which represents the goal of task $l$.
We then define the transition probabilities $\mathbb{P}(s_n^{t+1}|s_n^t,a_n^t)$ in accordance with the system uncertainties present in the corresponding edge.
Subsequently, we define a policy, $\pi_n:\calS_n\rightarrow\calA_n$, that maps states into actions.
With these definitions in place, the dynamic scheduling problem of edge $n$ can be formulated as a Markov decision process (MDP), expressed as follows:
\begin{equation}
	\label{eqn:mdp}
	\maximize_{\pi_n:\calS_n\to\calA_n} ~ U_{l(n)}^{\pi_n}(s_n) \triangleq \mathbb{E}\left[\left. \sum_{t=0}^{\infty}{\gamma^t u_{l(n)}(s_n^t,\pi_n(s_n^t))}\right|s_n^0=s_n \right],
\end{equation}
where $\gamma$ is a discount factor.
For this problem, the optimal value function can be defined by
\begin{equation}
	\label{eqn:mdp_optimal_value}
	J^*_n(s_n) = \max_{\pi_n} U^{\pi_n}_{l(n)}(s_n),~\forall s_n\in\calS_n,
\end{equation}
and its corresponding optimal policy is given by
\begin{equation}
	\label{eqn:mdp_optimal_policy}
	\pi^*_n = \argmax_{\pi_n} U^{\pi_n}_{l(n)}(s_n),~\forall s_n\in\calS_n.
\end{equation}
The problem formulation presented in \eqref{eqn:mdp} is widely used in the literature to represent a diverse range of dynamic scheduling tasks \cite{shani2005mdp, huang2021deep, lu2016partially, wei2018user, ye2018deep, xu2017deep, han2008resource, ferra2003applying, stidham1993survey, chang2000line}.
This popularity is due to the fact that most scheduling systems, including those in IoT networks, manage and identify their corresponding items/devices using indexing, as in the formulation.
In Appendix \ref{appendix:experiment}, we provide several examples of representative tasks in IoT networks.
These include wireless power transfer, data gathering, and radio resource scheduling, all of which is modeled using this formulation.

\section{Concept and Key Challenges on Collaborative Policy Learning for Dynamic Scheduling Tasks}
\label{sec:challenges}
In this section, we provide the concept of collaborative policy learning for dynamic scheduling tasks and discuss the key challenges involved in implementing it.

\subsection{Concept of Collaborative Policy Learning}
In the IoT network, each edge can optimally solve the dynamic scheduling problem in \eqref{eqn:mdp} by finding its optimal policy $\pi_n^*$ in \eqref{eqn:mdp_optimal_policy}.
To this end, standard dynamic programming (DP) approaches, such as value iteration and policy iteration, and traditional reinforcement learning (RL) approaches, such as SARSA and Q-learning, can be used.
However, DP approaches are generally impractical for practical applications, as they require perfect prior information on system uncertainties.
Also, both of DP and RL have a large computational complexity due to the curse of dimensionality.
To overcome these practical limitations, DRL has been widely used recently to solve such problems \cite{he2018green, shi2020deep, lee2019resource, peng2020learning, zhang2020hierarchical, lee2019contextual, xiong2020uav, lee2020adaptive}.
In DRL, an agent constructs a DNN that can represent the policy of the problem.
The agent then trains the DNN to approximate the optimal policy to solve the problem.
%
Consequently, each edge can solve its dynamic scheduling problem by training a policy represented by a DNN.

Since the policy is represented as a DNN with such an approach based on DRL, a central DNN (i.e. a central policy) for each task may be collaboratively learned at the cloud server.
To this end, we can use FRL to learn the central DNN by effectively aggregating the local DNNs (i.e., the local policies) from all edges conducting the task.
We now describe the FRL procedure for collaborative policy learning for dynamic scheduling tasks in the IoT network.
This process unfolds takes place over a discrete time horizon, which consists of multiple rounds denoted by $\calR=\{1,2,\ldots\}$.
The index of rounds is denoted by $r$.
The time horizon of FRL typically spans a larger time scale than that of each task.
As a result, FRL aggregates the DNNs, which are locally trained by the edges over multiple time slots, in each round.
Since we consider $L$ tasks, FRL is applied to $L$ DNNs in the cloud network.
The central parameters of the DNN for task $l$ at the cloud server are denoted by $\btheta_l$, and the local parameters of the DNN at edge $n$ are denoted by $\bw_n$.
We define the vector of the parameters of the DNNs of all edges as $\bW=(\bw_1,\ldots,\bw_N)$.
With these definitions, we can formally define the problem of the collaborative policy learning framework as follows:
\begin{equation}
\label{eqn:federated_learning_problem}
\minimize_{\bW}~ l(\bW) \triangleq \sum_{l\in\calL}\frac{1}{\bar{K}_l}\sum_{n\in\calN(l)}\sum_{k=1}^{K_n}f_n(\bw_n,k),
\end{equation}
where $K_n$ is the number of experiences from edge $n$, $\bar{K}_l=\sum_{n\in\calN(l)}K_n$, and $f_n(\bw_n,k)$ is an empirical loss function with $\bw_n$ at the $k$th experience of edge $n$.

To solve the problem, the cloud server broadcasts the central parameters, $\btheta_l^r$, for task $l$ in round $r$ to the edges in $\calN(l)$.
Then, in round $r$, each edge $n\in\calN(l)$ substitutes its local parameters, $\bw_n^r$, with $\btheta_l^r$.
After this substitution, each edge trains its local parameters using its local experiences.
These trained parameters are then uploaded to the cloud server.
The cloud server updates its central parameters for task $l$ by aggregating the received parameters from edges in $\calN(l)$, using
\begin{equation}
\label{eqn:federated_learning}
\btheta_l^{r+1}=\btheta_l^r-\sum_{n\in\calN(l)}c_n \nabla g_n^r,
\end{equation}
where $\nabla g_n^r$ is the local gradient of edge $n$ in round $r$, and $c_n$ is the central learning weight of edge $n$.
Here, $\nabla g_n^r$ reflects the disparity between the central parameter, $\btheta_l^r$, in round $r$ and the local parameters, ${\bw_n^r}'$, of edge $n$ following local training in round $r$.
Meanwhile, $c_n$ is established based on the contribution of edge $n$ to the central parameter updates for task $l(n)$, defined as
\begin{equation}
\label{eqn:federated_learning_weight}
c_n=\frac{K_n}{\sum_{n'\in\calN(l(n))}K_{n'}}.
\end{equation}
Once the central parameters are updated, the current round is completed.
The process then proceeds to the next round.
By repeating this process, FRL solves the problem in \eqref{eqn:federated_learning_problem}.

\subsection{Key Challenges on Collaborative Policy Learning}
\subsubsection{Limited Cloud Resources for Collaborative Policy Learning on Multiple Tasks}
\label{sec:challenge_task_selection}
FRL operates in an IoT network to handle multiple tasks.
However, as described in Section \ref{sec:model_iot_net}, it must do so using only limited cloud resources, such as computing power, memory, and network bandwidth.
This limitation implies that if there are not enough cloud resources to proceed with FRL for all tasks in each round, only a subset of tasks may be selected for FRL.
Specifically, the amount of cloud resources required to conduct FRL for each task depends on the number of edges participating in FRL.
In each round, some edges may be unable to participate in FRL due to various reasons, such as other higher-priority jobs or network shutdowns for energy-saving purposes.
However, in typical FRL, once the central parameters are updated by aggregating the local parameters of participating edges, the local parameters of non-participating edges are abandoned.
All edges' local parameters are then substituted by the central parameters.
This is because using outdated local parameters in FRL may negatively affect the convergence of central parameters \cite{lee2021adaptive}.\looseness=-1

Additionally, according to the convergence analysis of FRL, the effectiveness of FRL improves as the number of participants in FRL and corresponding data increases \cite{xia2020multi,lee2021adaptive}.
This implies that even if tasks are selected uniformly, the effectiveness of FRL for each task may vary significantly based on the number of participants.
Therefore, to ensure that all tasks benefit fairly from FRL, it is essential to consider fairness in terms of the number of participants, rather than the number of times they are selected.
In conclusion, to effectively utilize cloud resources for FRL, tasks should be carefully selected to maximize the number of participants while maintaining fairness among tasks in terms of the number of participants.
This issue will be addressed in Section \ref{sec:task_selection}.

\subsubsection{Collaborative Learning-Inapplicability of Conventional Policy Structures}
\label{sec:challenge_policy}
In this subsection, we explain why the conventional policy structures inapplicable to FRL for collaborative policy learning.
From the problem in \eqref{eqn:mdp}, we can see that the edges with task $l$ share an identical problem structure, which is defined by the state and decisions relevant to $K(l)$ and $G(l)$, respectively, and the reward function $u_l(s,a)$.
Accordingly, it seems feasible to collaboratively learn the policy for the task $l$ by using FRL (i.e., simply aggregating the DNNs from which edges with identical tasks locally train via DRL).
However, in practice, it is challenging to adopt FRL if DRL is directly applied to solve the problem in \eqref{eqn:mdp}, as in conventional works \cite{he2018green, shi2020deep, peng2020learning, zhang2020hierarchical}.
This is because the problems have different dynamics due to the varying number of IoT devices (i.e., $M_n$) and the transition probabilities.
For example, for edges $n_1$ and $n_2$ with varying numbers of IoT devices and system uncertainties, their state and action spaces can be different (i.e., $\calS_{n_1}\neq\calS_{n_2}$ and $\calA_{n_1}\neq\calA_{n_2}$), and their transition probabilities differ as well.
This implies that the DNNs for the policies in edges $n_1$ and $n_2$, based on the conventional works, have different structures (e.g., the DNNs may have different numbers of input and output units).
Besides, even though the state and action spaces are identical, they cannot be simply aggregated via FRL due to the different underlying statistical characteristics on the edges.
%
Therefore, one of the key challenges is that conventional dynamic scheduling policies are inapplicable to collaborative policy learning.
To overcome this issue, we need a policy structure that has a generalization capability over different edges, which implies that the policy for $l$ learned from one edge can be used other edges in $\calN(l)$.
Hence, such a policy structure allows us to collaboratively learn the DNN (i.e. a central policy) for task $l$ at the cloud server by effectively using the DNNs (i.e., the local policies) from all edges in $\calN(l)$.
%
%
This issue will be addressed in Section \ref{sec:edge-agnostic_policy}.

%
%

\section{Collaborative Policy Learning for Dynamic Scheduling Tasks in IoT Networks}
\label{sec:FL_framework}

In this section, we introduce two key enablers of collaborative policy learning for dynamic scheduling tasks in IoT networks.
First, we present a task selection algorithm tailored for efficient FRL in resource-limited IoT networks.
Second, we propose a policy structure suitable for collaborative learning in dynamic scheduling tasks.
These enablers address the key challenges outlined in Section \ref{sec:challenges}, laying the groundwork for a collaborative policy learning framework for dynamic scheduling tasks in IoT networks leveraging FRL.

\subsection{Opportunistic Task Selection for Effective Collaborative Policy Learning}
\label{sec:task_selection}
In this subsection, we address the issue of FRL for multiple tasks due to limited cloud resources raised in Section \ref{sec:challenge_task_selection}.
Firstly, we define the required resources for each participant (i.e., edge) with task $l$ as the required network bandwidth $B_l$, the required memory resources $O_l$, and the required computing resources $C_l$.
We then model the availability of each edge to participate in FRL in each round as a stationary process.
To represent the availabilities of all edges concisely, we define an availability state that corresponds to a combination of the availability conditions of all edges in a round and denote it by $p\in\calP$, where $\calP$ is the availability state space.
The availability indicator of edge $n$ in availability state $p$ is represented by $x_n^p\in\{0,1\}$, where $1$ indicates that edge $n$ is available to participate in FRL, and $0$ indicates that it is not.
The vector of the availability indicator of edges in availability state $p$ is defined as $\bx^p=(x_n^p)_{\forall n\in\calN}$.
The number of available edges with task $l$ in a round with availability state $p$ can be given as $x_l^p=\sum_{n\in\calN(l)}x_n^p$.
Then, the required bandwidth for task $l$ in a round with availability state $p$ is given by $B_l^p=x_l^pB_l$.
Similarly, the required memory resources and computing resources are given by $O_l^p=x_l^pO_l$ and $C_l^p=x_l^pC_l$, respectively.

For a task selection problem, we define a task selection indicator, $q_l^p$, for task $l$ in availability state $p$ as
\begin{equation}
	q_l^p = \begin{dcases}
		1, &\parbox[t]{.3\textwidth}{\textnormal{if task $l$ is selected for FRL in a round with availability state $p$,}}\\
		0, &\textnormal{otherwise.}
	\end{dcases}
\end{equation}
For convenience, we additionally define the vector of task selection indicators in availability state $p$ as $\bq^p=(q_l^p)_{\forall l\in\calL}$, and subsequently, the vector of all task selection indicators as $\bQ=(\bq^p)_{\forall p\in\calP}$.
Given that the required network bandwidth, memory resources, and computing resources for selected tasks must not exceed their corresponding maximum resources allowed for FRL in the cloud server, we consider the following constraints:
\begin{align}
	\sum_{l\in\calL}q_l^px_l^pB_l^p &\le B, ~ \forall p\in\calP, \label{eqn:task_selection_cost1} \\
	\sum_{l\in\calL}q_l^px_l^pO_l^p &\le O, ~ \forall p\in\calP, \label{eqn:task_selection_cost2} \\
	\sum_{l\in\calL}q_l^px_l^pC_l^p &\le C, ~ \forall p\in\calP. \label{eqn:task_selection_cost3}
\end{align}

As discussed in Section \ref{sec:challenge_task_selection}, effective FRL necessitates strategic task selection.
This strategy aims to maximize the number of participating edges in FRL and ensure that all tasks derive benefits.
To achieve this goal, we adopt a fairness concept in terms of the number of participating edges.
By taking the fairness into account in the average number of participating edges, we can guarantee that all tasks, including those operating at a smaller number of edges, benefit from FRL.
We calculate the average number of participants for task $l$ as $\sum_{p\in\calP}\phi^p q_l^p x_l^p$, where $\phi^p$ is the probability of the availability state being in $p$.
We then define the constraint of the minimum average number of participants for task $l$ as
\begin{equation}
	\label{eqn:task_selection_minimum_participants}
	\sum_{p\in\calP}\phi^p q_l^p x_l^p \geq X_l,~\forall l\in\calL,
\end{equation}
where $X_l$ is the required minimum average number of participants for task $l$.
Furthermore, we define the utility function for task $l$ as a function of its average number of participants, given by $V_l\left(\sum_{p\in\calP}\phi^p q_l^p x_l^p\right)$.
The utility function for task $l$ here is different from the reward function for task $l$ defined in Section \ref{sec:task_model}. The former is used in the task selection problem formulated in \eqref{eqn:task_selection_problem}, while the latter is used to represent the goal of task $l$.
Finally, we present the formulation of the task selection problem as follows:
\begin{IEEEeqnarray}{C'l}
	\label{eqn:task_selection_problem}
	\maximize_{\bQ} & \sum_{l\in\calL} V_l \left(\sum_{p\in\calP}\phi^p q_l^p x_l^p\right) \\
	\subjecto & \eqref{eqn:task_selection_cost1}, ~ \eqref{eqn:task_selection_cost2}, ~ \eqref{eqn:task_selection_cost3}, ~ \eqref{eqn:task_selection_minimum_participants}. \nonumber 
\end{IEEEeqnarray}
It is important to note that the task selection problem can accommodate various fairness definitions over tasks, such as proportional fairness and minmax fairness, by appropriately choosing the utility function and constraint parameters (i.e., $X_l$'s).
For example, we can achieve weighted proportional fairness if we choose the utility function as $V_l(\cdot)=w_l\log(\cdot)$, $\forall l\in\calL$, where $w_l$ is the weight of task $l$, and the constraint parameters as $X_l=-\infty$, $\forall l\in\calL$ (i.e., no constraints for the minimum average number of participants).
Hence, the choice of the utility function and constraint parameters can depend on the network characteristics, such as the size of DNNs and the complexity of tasks.\looseness=-1

We now develop an algorithm to optimally solve the task selection problem in \eqref{eqn:task_selection_problem}.
To this end, we first relax the integer variables into the continuous ones and introduce auxiliary variables $y_l$, $\forall l\in\calL$, which represent the average numbers of participants (i.e., $\sum_{p\in\calP} \phi^p q_l^p x_l^p$).
We then apply the Lagrangian approach and a stochastic subgradient algorithm, as in the opportunistic framework \cite{liu2003framework,kwon2010unified}.
Due to the page limit, the details of the task selection algorithm are omitted.
The task selection algorithm in round $r$ determines the task selection, $\bq^{(r)}=(q_l^{(r)})_{\forall l\in\calL}$, in round $r$, using\footnote{To explicitly denote the round, we use a superscript $(\cdot)^{(r)}$ instead of $(\cdot)^p$. This is justified because the availability state in each round is determined based on the system conditions in that specific round, such as the number of participants, the channel conditions, etc.}\looseness=-1
\begin{equation}
	\label{eqn:optimal_task_selection}
	\bq^{(r)}=\argmax_{(q_l)_{\forall l\in\calL}:\eqref{eqn:task_selection_cost1},\eqref{eqn:task_selection_cost2},\eqref{eqn:task_selection_cost3}} \left\lbrace \sum_{l\in\calL} \left(\lambda_l^{(r)}+\mu_l^{(r)}\right) q_l x_l^{(r)} \right\rbrace,
\end{equation}
where $\lambda_l^{(r)}$ is the Lagrange multiplier of task $l$ in round $r$ with respect to the auxiliary variable $y_l$, $\mu_l^{(r)}$ is the one with respect to the constraint in \eqref{eqn:task_selection_minimum_participants}, and $x_l^{(r)}$ is the number of available edges with task $l$ in round $r$.
At the end of round $r$, the Lagrange multipliers are updated, using
\begin{align}
	\lambda_l^{(r+1)} &= \left[\lambda_l^{(r)}-\alpha^{(r)} \left( q_l^{(r)}x_l^{(r)}-y_l^{(r)} \right) \right]^+, \label{eqn:lagrangian_update} \\
	\mu_l^{(r+1)} &= \left[\mu_l^{(r)}-\alpha^{(r)} \left( q_l^{(r)}x_l^{(r)}-X_l \right) \right]^+, \label{eqn:lagrangian_update2}
\end{align}
where $[\cdot]^+ = \max\{0,\cdot\}$, $\alpha^{(r)}$ is the positive step size in round $r$, and $y_l^{(r)}=\argmax_{y_l\geq 0}\lbrace V_l(y_l)-\lambda_l^{(r)}y_l \rbrace$.
We can demonstrate the optimality of this algorithm using the following theorem.

\begin{theorem}
	The task selection algorithm described in \eqref{eqn:optimal_task_selection}, \eqref{eqn:lagrangian_update}, and \eqref{eqn:lagrangian_update2} optimally solves the dynamic scheduling task selection problem in \eqref{eqn:task_selection_problem}.
\end{theorem}

In the interest of brevity, we refer readers to \cite{kwon2010unified} for the proof.
Moving forward, implementing the algorithm necessitates solving the task selection problem in \eqref{eqn:optimal_task_selection} for each round.
By denoting the weight of task $l$ in round $r$ as $w^{(r)}_l=(\lambda_l^{(r)}+\mu_l^{(r)})x_l^{(r)}$, we can recast the problem as
\begin{equation}
	\label{eqn:task_selection_problem_knapsack}
	\max_{(q_l)_{\forall l\in\calL}:\eqref{eqn:task_selection_cost1},\eqref{eqn:task_selection_cost2},\eqref{eqn:task_selection_cost3}} \sum_{l\in\calL} w_l^{(r)} q_l.
\end{equation}
We denote the solution to this problem as $\bq^{(r)}$.
Notably, the problem in \eqref{eqn:task_selection_problem_knapsack} is a typical multidimensional knapsack problem \cite{kellerer2004multidimensional}, which can be solved efficiently using dynamic programming or branch-and-bound methods \cite{kellerer2004multidimensional}.

\subsection{Collaborative Learning-Applicable Edge-Agnostic Policy Structure for General Dynamic Scheduling Tasks}
\label{sec:edge-agnostic_policy}
As we discussed in Section \ref{sec:task_model}, the diverse range of dynamic scheduling tasks, such as wireless power transfer, data gathering, and radio resource scheduling, have an identical problem structure in \eqref{eqn:mdp}.
Hence, if multiple edges share such identical tasks, for each task, they also share the identical problem structure, which is determined by the types of state information, decisions, and reward function for the task.
However, as emphasized in Section \ref{sec:challenge_policy}, the edges typically have different \textit{dynamics,} due to the varying number of IoT devices and system uncertainties.
This renders conventional dynamic scheduling policies to be inapplicable to collaborative policy learning because of its lack of generalization capability, as discussed in Section \ref{sec:challenge_policy}.

To ensure that a policy for each task is capable of generalization over different edges that conduct the task, it should be designed to represent states and actions in a way that is independent of the dynamics of the edges.
Furthermore, the policy should be able to learn a scheduling principle, capable of identifying which \textit{condition (i.e., state information)} of IoT devices is more favorable for effective scheduling.
For example, suppose a policy that represents such a principle for task $l$.
Then, any edge that conducts task $l$ could identify its best IoT device to schedule by comparing the current conditions of all IoT devices based on the policy.
If each edge learns a DNN that represents such a policy using DRL, the DNNs from all edges can be aggregated via FRL thanks to the generalization capability across the edges.
Consequently, it would enable collaborative policy learning.\looseness=-1

Here, we propose a collaborative learning-applicable edge-agnostic policy structure that satisfies the aforementioned features by borrowing the concept of the circumstance-independent (CI) policy structure in \cite{lee2019resource}.
The CI policy structure represents a policy for the radio resource scheduling problem in a single-cell wireless network, regardless of the network's dynamic circumstances.
For an edge-agnostic policy structure, we generalize the concept of the CI policy structure for dynamic scheduling tasks and extend it to be used in multiple edges for FRL.
Next, we present edge-agnostic state and action structures. These structures focus on the \textit{conditions} of the IoT devices in each edge, rather than on each \textit{IoT device} itself, as in \eqref{eqn:task_state_definition} and \eqref{eqn:task_action_definition}.

\subsubsection{Structure of Edge-Agnostic State, Action, and Policy}
\begin{figure}[!t]
	\centering
	\includegraphics[width=0.8\linewidth]{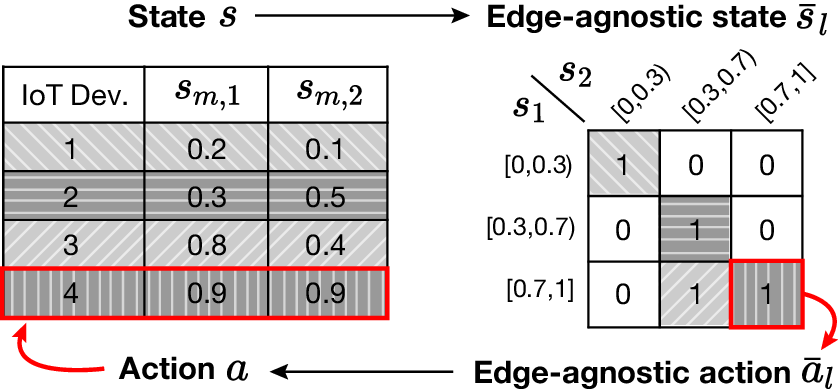}
	\caption{Illustration of edge-agnostic state and action.\label{fig:edge-agnostic_state}}
	\vspace{-1em}
\end{figure}

We first define an edge-agnostic state that represents the conditions of IoT devices in any edges.
Specifically, it indicates whether an IoT device with a specific state information condition exists or not in each time slot.
To achieve this, the space of each $k$th state information of task $l$, where $k\in\{1,2,\ldots,K(l)\}$, is partitioned into $H_{k,l}$ disjoint intervals.
The intervals in the partitions for $k$th state information are indexed by $h_{k,l}\in\{1,2,\ldots,H_{k,l}\}$.
The condition of each IoT device in the edge with task $l$ can then be represented as a combination of the intervals of each state information, as illustrated in Fig. \ref{fig:edge-agnostic_state}.
The structure of the edge-agnostic state for task $l$ is defined as a $K$-dimensional matrix whose size is given by $\prod_{k\in\{1,\ldots,K(l)\}}H_{k,l}$.
Each element of the state is indexed by a tuple $h=(h_{1,l},\ldots,h_{K(l),l})$.
Formally, we denote the edge-agnostic state for task $l$ by $\bar{s}_l$ and define it as\looseness=-1
\begin{equation}
	\label{eqn:edge-agnostic_state}
	\bar{s}_l(h) = \begin{dcases}
		1, &\textnormal{if there exists any IoT device in condition $h$}, \\
		0, &\textnormal{otherwise},
	\end{dcases}
\end{equation}
where $\bar{s}_l(h)$ denotes the element of state $\bar{s}_l$ whose index is given by $h$.
The edge-agnostic state space for task $l$ can be defined by $\bar{\calS}_l=\{0,1\}^{\prod_{k\in\{1,\ldots,K(l)\}}H_{k,l}}$.
It is noteworthy that the edge-agnostic state for each task can describe the conditions of IoT devices in any edges with the task, regardless of the number of IoT devices.

Based on the edge-agnostic state, we can define an edge-agnostic action that indicates the \textit{condition} to be scheduled rather than a specific IoT device.
Specifically, the edge-agnostic action for task $l$ can be defined using the index of the element of the edge-agnostic state and relevant scheduling decisions as
\begin{equation}
	\label{eqn:edge-agnostic_action}
	\bar{a}_l =(h_{1,l},\ldots,h_{K(l),l},g_{1,l},\ldots,g_{G(l),l})\in\bar{\calA}_l,
\end{equation}
where $\bar{\calA}_l$ is the edge-agnostic action space for task $l$.
Note that not all combinations of conditions in the edge-agnostic state may be feasible for scheduling, as there may not be any IoT device satisfying a particular condition.
Thus, we define the feasible edge-agnostic action space with state $\bar{s}_l$ as\looseness=-1
\begin{equation}
	\bar{\calA}_l(\bar{s}_l)=\{\bar{a}_l\in\bar{\mathcal{A}}_l| \bar{s}_l(h) = 1\}.
\end{equation}

With the aforementioned elements, an edge-agnostic policy for task $l$ can be defined as $\bar{\pi}_l:\bar{\calS}_l\rightarrow\bar{\calA}_l$.
As the general scheduling principle, the edge-agnostic state and action for task $l$ focus on the \textit{condition} of IoT devices, and the edge-agnostic policy represents the selection of specific \textit{condition} in scheduling, rather than the selection of the index of a specific IoT device.
This implies that when the edge-agnostic policy for each task is learned through DRL, its corresponding DNN is trained to approximate the optimal general scheduling principle for the task that has a generalization capability over different edges.
Therefore, the edge-agnostic policy can be utilized in any edges with task $l$, even if the edges have different dynamics such as the numbers of IoT devices.
Consequently, FRL can be applied to the DNN for collaborative learning thanks to the generalization capability.

\subsubsection{DRL for Learning Edge-Agnostic Policy}
\label{sec:drl_policy}
We propose a procedure to learning the edge-agnostic policy via DRL in edge $n$, based on the system model described in Section \ref{sec:task_model}.
Specifically, using the reward function $u_{l(n)}$ defined in the section, edge $n$ can learn the edge-agnostic policy $\pi_{l(n)}$ that solves the dynamic scheduling problem of edge $n$ (and can also be used for other edges with task $l(n)$) via DRL methods.
For the sake of simplicity, we describe the proposed approach based on the well-known deep Q-network (DQN) in \cite{mnih2015human}, but other methods can also be employed.\looseness=-1

In DQN, a DNN is employed to approximate the optimal action-value function, based on the edge-agnostic states and actions.
The DNN structure for the edge-agnostic policy is determined based on task $l(n)$, and all edges associated with the same task have an identical DNN structure.
We denote the parameters of the DNN for task $l$ by $\btheta_l$ and those in edge $n$ by $\bw_n$.
Consequently, the DNN in edge $n$, $\bw_n$, has a structure identical to that of $\btheta_l(n)$.
The optimal action-value function with a given $\bar{s}_{l(n)}$ and $\bar{a}_{l(n)}$ is denoted by $\bar{Q}^*_{l(n)}(\bar{s}_{l(n)},\bar{a}_{l(n)})$, while its Q-approximation derived from the DNN is denoted by $\bar{Q}_{l(n)}(\bar{s}_{l(n)},\bar{a}_{l(n)};\bw_n)$.
In time slot $t$, the observed state $s_n^t$ in accordance with \eqref{eqn:task_state_definition} is translated into the edge-agnostic state $\bar{s}_{l(n)}^t$ as per \eqref{eqn:edge-agnostic_state}.
Based on $\bar{s}_{l(n)}^t$, the edge-agnostic policy chooses the edge-agnostic action $\bar{a}_{l(n)}^t$ from $\bar{\calA}_l(\bar{s}_l^t)$, according to its exploration-exploitation strategy (for instance, an $\epsilon$-greedy method).
Subsequently, the selected edge-agnostic action $\bar{a}_{l(n)}^t$ in line with \eqref{eqn:edge-agnostic_action} is translated into the action $a_n^t$ as per \eqref{eqn:task_action_definition}.
When more than one IoT device fulfills the condition indicated by the edge-agnostic action, one of these IoT devices is arbitrary selected as the scheduled IoT device for that time slot.
The translation of states and actions is illustrated in Fig. \ref{fig:edge-agnostic_state}.
After scheduling, the reward $u_{l(n)}^t(s_n^t,a_n^t)$ and the next state $s_n^{t+1}$ are observed.
Then, an edge-agnostic experience sample for time slot $t$ is generated as $(\bar{s}_{l(n)}^t,\bar{a}_{l(n)}^t,u_{l(n)}^t,\bar{s}_{l(n)}^{t+1})$.
Using these experience samples, the DNN is trained in line with standard DQN methods, incorporating experience replay and fixed-target Q-network.

\subsection{Collaborative Policy Learning Framework for Dynamic Scheduling Tasks Using FRL}
In this subsection, we propose a collaborative policy learning framework for dynamic scheduling tasks in IoT networks leveraging FRL.
The framework is built upon the task selection algorithm and the collaborative learning-applicable scheduling policy discussed in previous subsections.
Initially, both the cloud server and each edge initialize their DNNs to learn the edge-agnostic policy applicable to dynamic scheduling tasks.
In the cloud server, the central parameters of the DNNs for all tasks are initialized as $\btheta_l^1$, $\forall l\in\calL$, to facilitate FRL across all tasks.
Concurrently, each edge $n$ initializes the local parameters, $\bw_n$, of its DNN as $\btheta_{l(n)}^1$ to maintain identical DNN structures for the same task.
Moreover, the edge sets its local parameters, $\bw_n^1$, at the onset of the first round to be $\bw_n$.
It is crucial to understand that $\bw_n$, without the round index, represents the local parameters trained in the DQN algorithm at edge $n$.
Subsequently, edge $n$ begins executing its DQN algorithm with its local parameters, $\bw_n$, as described in Section \ref{sec:drl_policy}.
Notably, these DQN algorithms operate concurrently and can be temporarily suspended to accommodate FRL.

\begin{algorithm}[t]
\caption{Procedure of Collaborative Policy Learning Framework for Dynamic Scheduling Tasks}\label{alg:framework}
\footnotesize
\begin{algorithmic}[1]
\State The cloud server initializes DNN $\btheta_l^1$, $\forall l\in\calL$, and $\blambda^{(1)}$.
\State Edge $n$ initializes DNN $\bw_n$ as $\btheta_{l(n)}^1$ and sets $\bw_n^1$ to be $\bw_n$, $\forall n\in\calN$.
\State Edge $n$ starts to run \textsc{DQN}($\bw_n$) individually, $\forall n\in\calN$.
\For{round $r\in\calR\triangleq\{1,2,\ldots\}$}
\State The cloud server observes $\bx^{(r)}$.
\State The cloud server obtains the task selection decision $\bq^{(r)}$ using \eqref{eqn:optimal_task_selection}.
\For{task $l\in\{l':q_{l'}^{(r)}=1\}$}\Comment{in parallel}
\State The cloud server runs \textsc{FedDS}($l,\bx^{(r)}$).
\EndFor
\For{task $l\notin\{l':q_{l'}^{(r)}=1\}$}\Comment{in parallel}
\State The cloud server sets $\btheta_l^{r+1}$ to be $\btheta_l^r$.
\State Edge $n\in\calN(l)$ sets $\bw_n^{r+1}$ to be $\bw_n^r$.
\EndFor
\State The cloud server updates its Lagrange multipliers using \eqref{eqn:lagrangian_update} and \eqref{eqn:lagrangian_update2}.
\EndFor
\Procedure{DQN}{$\bw_n$} \Comment{at edge $n$}
\State Observe $s_n^1$ and translate it into $\bar{s}_{l(n)}^1$.
\For{time slot $t\in\{1,2,\ldots\}$}
\State Choose $\bar{a}_{l(n)}^t\in\bar{\calA}(\bar{s}_{l(n)}^t)$ and translate action $\bar{a}_{l(n)}^t$ into $a_n^t$.
\State Do $a_n^t$, observe $u_{l(n)}^t$ and $s_n^{t+1}$, and translate state $s_n^{t+1}$ into $\bar{s}_{l(n)}^{t+1}$.
\State Store experience $\left(\bar{s}_{l(n)}^t,\bar{a}_{l(n)}^t,u_{l(n)}^t,\bar{s}_{l(n)}^{t+1}\right)$.
\State Update $\btheta_n$ using its experiences to learn the target Q-value.
\EndFor
\EndProcedure
\Procedure{FedDS}{$l,\bx^{(r)}$}
\For{edge $n\in\{n':n'\in\calN(l)\textrm{ and }x_{n'}^{(r)}=1\}$}
\Comment{in parallel}
\State Edge $n$ temporarily pauses its \textsc{DQN}.
\State Edge $n$ stores the current DNN into $\bw_n$.
\State{Edge $n$ calculates the local gradients $\nabla g_n^r$ from the DNN $\bw_n^r$ to the current one $\bw_n$.}
\State Edge $n$ uploads $\nabla g_n^r$ to the cloud server.
\EndFor
\State{The cloud server calculates the DNN $\btheta_l^{r+1}$ by aggregating the local gradients using \eqref{eqn:federated_learning_2}}.
\State{The cloud server broadcasts $\btheta_l^{r+1}$ to all edges in $\calN(l)$.}
\State{Edge $n\in\calN(l)$ replaces its DNN $\bw_n$ with $\btheta_l^{r+1}$ and sets $\bw^{r+1}_n$ to be $\bw_n$.}
\State Every edge with the paused DQN resumes its DQN.
\EndProcedure
\end{algorithmic}
\end{algorithm}

During round $r$, the cloud server evaluates the availability of the edges to engage in FRL, denoted as $\bx^{(r)}$.
Based on this assessment, it makes a task selection decision, denoted as $\bq^{(r)}$, in accordance with \eqref{eqn:optimal_task_selection}.
This selection ensures the convergence of FRL for tasks, as we will demonstrate later.
For each selected task $l$, the cloud server and the available edges conduct FRL for the task through \textsc{FedDS} in parallel.
During this process, every available edge $n$ associated with task $l$ (i.e., $n\in\{n':n'\in\calN(l)\textnormal{ and }x_{n'}^{(r)}=1\}$) temporarily suspends its DQN algorithm to maintain the current local parameters $\bw_n$.
Edge $n$ calculates the local gradients, $\nabla g_n^r$, utilizing the local parameters of its DNN at the onset of round $r$, denoted as $\bw_n^r$, and the current ones, denoted as $\bw_n$.
Following this, edge $n$ uploads the local gradients to the cloud server.
Upon receiving the local gradients from the available edges during round $r$, the cloud server computes the central parameters of the DNN for task $l$, denoted as $\btheta_l^{r+1}$, using
\begin{equation}
\label{eqn:federated_learning_2}
\btheta_l^{r+1}=\btheta_l^r-\sum_{n\in\calN(l)}c_n^rx_n^{(r)}\nabla g_n^r,
\end{equation}
where $c_n^r=N_lc_n/x_l^{(r)}$ with $N_l$ being the number of edges associated with task $l$.
This procedure trains the edge-agnostic policy for task $l$ by gathering experiences from all available edges associated with task $l$.
Following this, the cloud server broadcasts the updated central parameters, $\btheta_l^{r+1}$, to all edges with task $l$.
Each edge $n$ with task $l$ substitutes its locally trained parameters from its DQN algorithm with $\btheta_l^{r+1}$.
It then sets its local parameters at the onset of round $r+1$, denoted as $\bw_n^{r+1}$, to be $w_n$.
Once each edge resumes its previously paused DQN algorithm, the FRL process concludes.
Subsequent to the FRL process, the cloud server updates the Lagrange multipliers for tasks as depicted in \eqref{eqn:lagrangian_update} and \eqref{eqn:lagrangian_update2} to ensure fairness across them, as defined in relation to the task selection problem.
The entire framework is summarized in Algorithm \ref{alg:framework}.

\subsection{Convergence Analysis of Collaboartive Policy Learning}
In this subsection, we provide a convergence analysis of the proposed collaborative policy learning framework.
To this end, we first introduce the following assumptions which are typical ones in the literature \cite{ruan2021towards}:

\begin{assumption}
	The objective function of FL $l(\bw)$ is $L$-smooth, which means it has a Lipschitz continuous gradient with a constant $L>0$. Symbolically, this can be written as, for any two points $\bw_1$ and $\bw_2$, $l(\bw_1)-l(\bw_2)\leq \langle\nabla l(\bw_2),\bw_1-\bw_2 \rangle+\frac{L}{2}\Vert\bw_1-\bw_2\Vert^2$.
\end{assumption}

\begin{assumption}
	The objective function of FL $l(\bw)$ is $\xi$-strongly convex with $\xi>0$, which means that for any $\bw_1$ and $\bw_2$, the following inequality holds: $l(\bw_1)-l(\bw_2)\geq \langle\nabla l(\bw_2),\bw_1-\bw_2 \rangle+\frac{\xi}{2}\Vert\bw_1-\bw_2\Vert^2$.
\end{assumption}

\begin{assumption}
	The variance of the gradients at each edge is bounded for all rounds, i.e., $\mathbb{E}\Vert g_n^r-\bar{g}_n^r\Vert^2\leq V^2$, $\forall n, r$, where $\bar{g}_n^r$ denotes the mean of the gradients at edge $n$ in round $r$.
\end{assumption}

\begin{assumption}
	The expected squared norm of the gradients at each edge is uniformly bounded for all rounds, i.e., $\mathbb{E}\Vert g_n^r\Vert^2\leq V^2$, $\forall n, r$.
\end{assumption}

To capture and quantify the non-independent and identically distributed (non-i.i.d.) experiences among edges, we introduce a parameter to represent the degree of experience distribution difference for edge $n$, expressed as $\Gamma_{l(n)}^n=f_n(\bw_{l(n)}^*)-f_n^*$, where $\bw_l^*$ denotes the minimizer of the loss function for task $l$, and $f_n^*$ represents the minimum value of $f_n$.
Subsequently, we define $\Gamma_l=\sum_{n\in\calN(l)}c_n\Gamma_l^n$.
We proceed under the assumption that each edge participates in FRL during each round with equal probability.
Given this assumption, we can demonstrate the convergence of FRL for dynamic scheduling tasks using the forthcoming theorem.

\begin{theorem}
\label{thm:convergence}
	The collaborative policy learning framework for dynamic scheduling tasks in Algorithm \ref{alg:framework} achieves the following convergence rate of the target DNN for task $l$:
\begin{equation}
	O\left(\left({N_l}^2+\bar{\sigma}_l^2+\Gamma_l\right)T^{-1}\right),
\end{equation}
where $T$ is the number of rounds, and $\bar{\sigma}_l^2=\sum_{n\in\calN(l)}(c_n\sigma_n)^2$.
\end{theorem}
\begin{IEEEproof}
See Appendix \ref{appendix:proof}.
\end{IEEEproof}

Theorem \ref{thm:convergence} takes consideration of the opportunistic task selection in Section \ref{sec:task_selection} compared with the analysis in \cite{ruan2021towards}.
Consequently, it clearly shows the convergence of the proposed collaborative policy learning framework.

\section{Experimental Results}
\label{sec:experimental_results}
In this section, we showcase experimental results evaluating the performance of our proposed collaborative policy learning framework for dynamic scheduling tasks.
To achieve this, we have created a dedicated Python-based simulator and run simulations on a simulated IoT network composed of multiple edges.
Each edge is assigned one of the following three tasks:
\begin{itemize}
\item \textbf{Task A: Wireless power transfer task} \textendash{} This task aims to minimize power outages of IoT devices attributable to low battery levels \cite{lee2019contextual}.
In each time slot, an AP wirelessly transfers power to a selected IoT device, with the charging rate being dependent on the device's channel condition.
If an IoT device is in an active state, its battery is discharged at a given rate.
The active state stochastically changes based on a Markov model.
The battery level of each IoT device is updated according to its active state and the amount of wireless power transferred from the AP.
The active state, battery level, and charging rate of each IoT device are used as state information.
The cost in each time slot is determined by the number of IoT devices whose battery level is below a threshold and those whose battery is empty.
The negative cost is taken as a reward.
\item \textbf{Task B: Data gathering task} \textendash{} This task aims to maximize the number of gathered data samples while minimizing dropped data samples in an IoT network \cite{kaur2020data}.
In each time slot, an AP selects an IoT device to transmit its data samples to the AP.
The transmission capacity of each IoT device to gather the data sample is time-varying, and the data samples randomly arrive at the buffer of each IoT device.
If the buffer overflows, the exceeded data samples are dropped.
The remaining buffer size and transmission capacity of each IoT device are used as state information.
The reward in each time slot is defined as the the number of gathered data samples minus the number of dropped data samples in that time slot.
\item \textbf{Task C: Radio resource scheduling task} \textendash{} This task aims to minimize the transmission power at an AP while ensuring the minimum average data rate requirements of IoT devices \cite{lee2019resource}.
In each time slot, an AP selects an IoT device to serve and the corresponding transmission power.
The determined transmission power then impacts the achievable data rate for the IoT device, following the Shannon capacity. 
The data rate depends on the IoT device's channel gain, which varies over time based on a channel model with a log-normal shadowing.
IoT devices also update their degree of dissatisfaction regarding the data rate requirements (DoD).
The channel gain and DoD of each IoT device are used as state information.
The reward for each time slot is calculated as the achieved data rate weighted by the DoD minus the transmission power.
\end{itemize}
We consider three distinct scenarios for each task to demonstrate the edge-agnostic feature.
Take the radio resource scheduling task as an example, where we examine three different scenarios with varying numbers of users and data rate requirements.
For a more comprehensive understanding of each task, please refer to Appendix \ref{appendix:experiment}.
Furthermore, to facilitate comparative analysis of each task's performance, we normalize the reward in the subsequent results.

We now present the basic simulation settings, which form a base setup and are consistently applied unless otherwise specified.
For each of the three tasks, we consider a total of twenty edges: seven for scenario A, seven for scenario B, and six for scenario C, which altogether constitute sixty edges in the IoT network.
We set the arrival rates for the edges with tasks A, B, and C to $0.7$, $0.4$, and $0.4$, respectively.
The maximum network bandwidth, memory, and computing resources of the IoT networks for federated learning are all set to $21$.
The DQN algorithm employs a fully-connected DNN with three hidden layers of $300$ units across all tasks.
We set the learning rate, batch size, train interval, and target update interval to $10^{-5}$, $32$, $50$, and $100$, respectively.
Given that all tasks share the same DNN structure, the bandwidth, memory, and computing resources required at each edge are identical for all tasks.
Consequently, without loss of generality, we assign a ujnit value to these parameters across all tasks.
The number of time slots per round for federated learning is set to $250$, and the simulation is run over $200$ rounds.

To assess the performance of our collaborative policy learning framework, we compare it to both an ideal benchmark and a baseline that excludes FRL. The algorithms utilized in this comparison are defined as follows:
\begin{itemize}
\item \textbf{Bench} represents an ideal benchmark algorithm that is founded on our framework, but it neglects the maximum resource constraints for FRL as indicated in \eqref{eqn:task_selection_cost1}, \eqref{eqn:task_selection_cost2}, and \eqref{eqn:task_selection_cost3}.
This is a theoretical model and cannot be practically achieved.
In each round, Bench always conducts FRL for all tasks, thereby delivering a performance upper bound.

\item \textbf{FL-PF} represents our framework with a proportional fair task selection. It is implemented by setting the utility function of tasks to a logarithm function (i.e., $V_l(x)=\log(x)$, $\forall l\in\calL$), and the required minimum average number of participants for task $l$ to $5$ (i.e., $X_l=5$).

\item \textbf{FL-Greedy} represents our framework with a greedy task selection.
In each round, tasks are selected as much as possible in a decreasing order of the number of available edges.

\item \textbf{FL-RR} represents our framework that employs a round-robin task selection.
In each round, tasks are selected as much as possible in a round-robin way (i.e., in a circular order of tasks).\looseness=-1

\item \textbf{No-FL} represents an algorithm without FRL.
In this algorithm, each edge learns its policy independently and individually.
This is implemented by setting the task selection indicators for all tasks and rounds to $0$ (i.e.,  $q_l^r=0$, $\forall l\in\calL$, $\forall r\in\calR$).
\end{itemize}

\subsection{Participants of Collaborative Policy Learning}
\label{sec:participants_FL}

\begin{figure}[!t]
	\centering
	\includegraphics[width=0.7\linewidth]{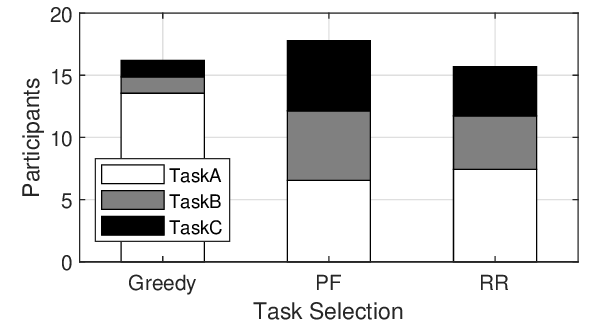}
		\vspace{-0.1in}
	\caption{Sum of the average number of participating edges for each task.\label{fig:tot_par}}
\end{figure}

\begin{figure*}[!t]
\vspace{-0.1in}
	\centering
	\includegraphics[width=0.6\linewidth]{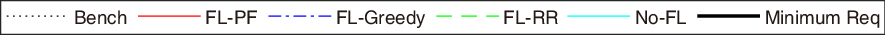}
	
	\subfloat[Task A.]{\includegraphics[width=0.32\linewidth]{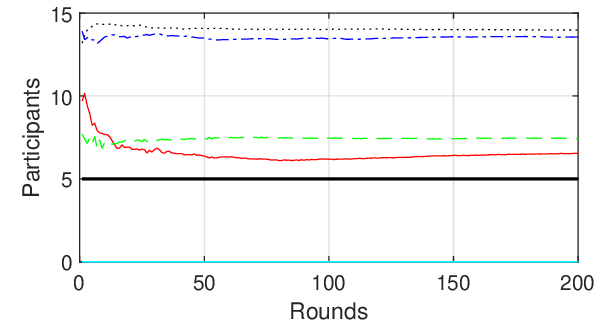}
		\label{fig:taskA_participants}}
	\hfil
	\subfloat[Task B.]{\includegraphics[width=0.32\linewidth]{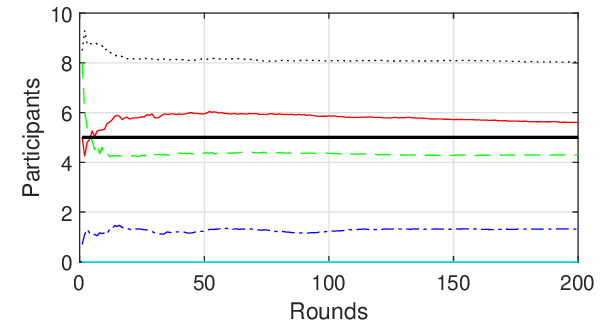}
		\label{fig:taskB_participants}}
	\hfil
	\subfloat[Task C.]{\includegraphics[width=0.32\linewidth]{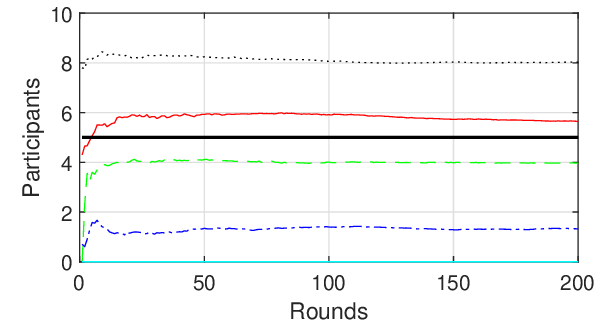}
		\label{fig:taskC_participants}}

	\caption{Average number of participants for each task.}
	\label{fig:participants}
\end{figure*}

We first present the sum of the average numbers of participants (i.e., participating edges) for all tasks in Fig. \ref{fig:tot_par}.
Note that Bench and No-FL are excluded from the figure as Bench reaches the maximum number of participants without resource constraint, while No-FL consistently achieves zero participants.
As observed from the figure, FL-PF attains a greater total number of participants than Greedy and RR, while ensuring fairness among the tasks in terms of participant numbers.
Conversely, FL-Greedy and FL-RR lead to more edges participating in task A than tasks B and C.
This imbalance creates unfairness among the tasks and may result in tasks B and C not achieving enough performance improvement from collaborative policy learning.
These observations suggest that FL-PF selects tasks in a manner that promotes effective collaborative policy learning, considering the time-varying availability conditions of edges and limited resources.
We delve into a more detailed comparison of the performances of the different algorithms in the following subsections.\looseness=-1

To illustrate the achievement of the minimum average number of participants, we depict the average number of participants for each task in Fig. \ref{fig:participants}.
As shown in Fig. \ref{fig:taskA_participants}, all collaborative policy learning algorithms (i.e., Bench, FL-PF, FL-Greedy, and FL-RR) successfully meet the minimum number of participants requirement.
Due to task A having the highest arrival rate, FL-Greedy excessively selects task A in nearly every round, which results in a participant count close to that of Bench.
However, as indicated in Figs. \ref{fig:taskB_participants} and \ref{fig:taskC_participants}, only FL-PF fulfills the minimum number of participants for tasks B and C.
FL-Greedy falls short of the minimum because of its skewed selection towards task A.
Conversely, while FL-RR selects tasks in a circularly fair manner, it does not take the number of participants into account, leading to fluctuating participant counts that depend on the arrival rate of each task.
From these figures, it is clear that FL-PF consistently meets the minimum number of participants across all tasks.

\subsection{Rewards of Dynamic Scheduling Tasks}
\label{sec:rewards_tasks}

\begin{figure}[!t]
	\centering
	\includegraphics[width=0.7\linewidth]{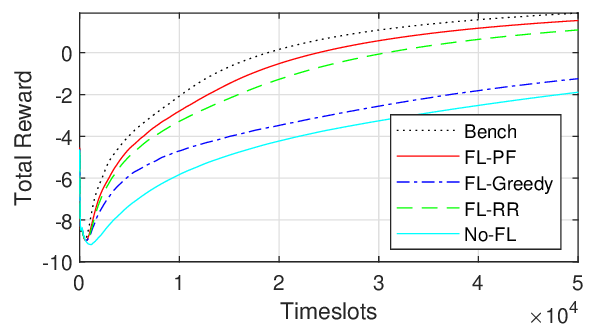}
	\caption{Sum of the average rewards of all edges.\label{fig:tot_rew}}
	\vspace{-0.1in}
\end{figure}

In Fig. \ref{fig:tot_rew}, we provide the sum of the average rewards of all edges.
As observed from the figure, all collaborative policy learning algorithms exhibit superior performance compared to No-FL.
Notably, FL-PF outperforms FL-RR and FL-Greedy and closely matches the performance of Bench.
This evidently demonstrates that FL-PF selects tasks in a more effective manner for collaborative policy learning compared to FL-RR and FL-Greedy.\looseness=-1

\begin{figure*}[!t]
\vspace{-0.1in}
	\centering
	\subfloat[Task A.]{\includegraphics[width=0.32\linewidth]{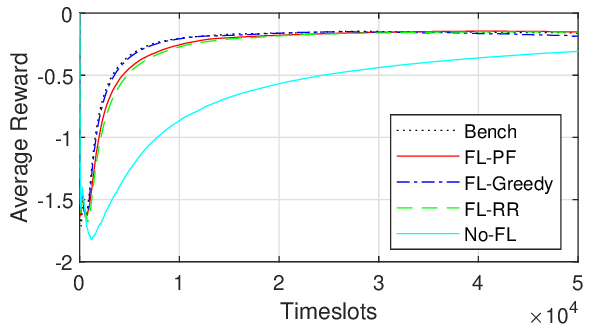}
		\label{fig:rew_t1}}
	\hfil
	\subfloat[Task B.]{\includegraphics[width=0.32\linewidth]{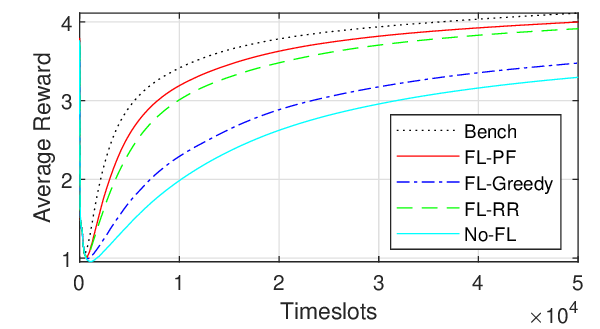}
		\label{fig:rew_t2}}
	\hfil
	\subfloat[Task C.]{\includegraphics[width=0.32\linewidth]{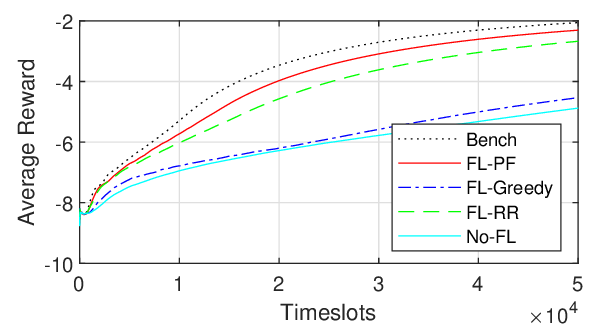}
		\label{fig:rew_t3}}
	\hfil

	\caption{Average rewards of the edges with each task.}
	\label{fig:rewards}
\end{figure*}

To delve deeper, we present the average reward of edges for each task in Fig. \ref{fig:rewards}.
Fig. \ref{fig:rew_t1} reveals that all collaborative policy learning algorithms yield similar rewards, significantly exceeding that of No-FL.
Interestingly, FL-Greedy secures an average reward almost identical to that of Bench.
In Figs. \ref{fig:rew_t2} and \ref{fig:rew_t3}, FL-PF surpasses both FL-RR and FL-Greedy, attaining a reward close to Bench.
While FL-Greedy only marginally outperforms No-FL, the other collaborative policy learning algorithms significantly surpass it.

Fig. \ref{fig:rewards} also reflects the relationship between the performance of collaborative policy learning and the number of participants, as demonstrated in Fig. \ref{fig:participants}.
From the figures, it is clear that the number of participants and the reward follow similar trends.
For tasks B and C, FL-PF reaps larger rewards compared to FL-RR, while also achieving a higher number of participants.
Moreover, FL-Greedy secures rewards nearly equal to Bench for task A, which boasts a large number of participants.
Conversely, it achieves rewards comparable to No-FL for tasks B and C, which have very few participants.
These findings strongly suggest that fairness among tasks, in terms of the number of participants, should be considered to ensure performance improvement from collaborative policy learning across all tasks.


\begin{figure*}[!t]
\vspace{-0.1in}
	\centering
	\subfloat[Task A.]{\includegraphics[width=0.32\linewidth]{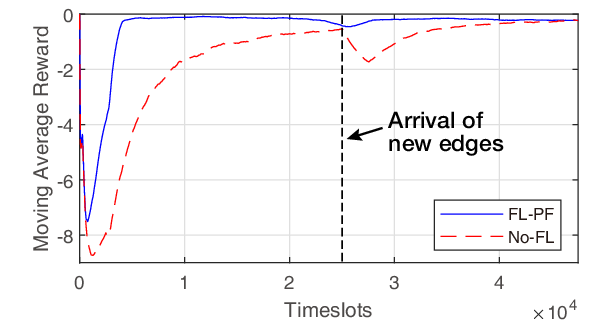}
		\label{fig:moving_avg_t1}}
	\hfil
	\subfloat[Task B.]{\includegraphics[width=0.32\linewidth]{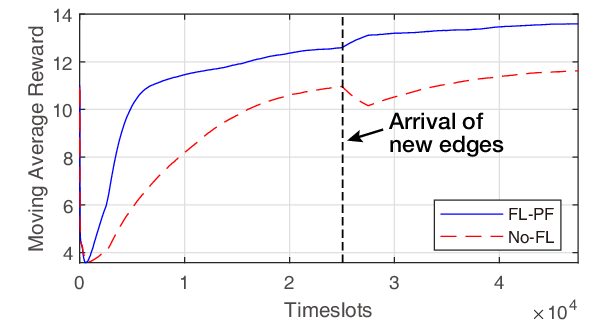}
		\label{fig:moving_avg_t2}}
	\hfil
	\subfloat[Task C.]{\includegraphics[width=0.32\linewidth]{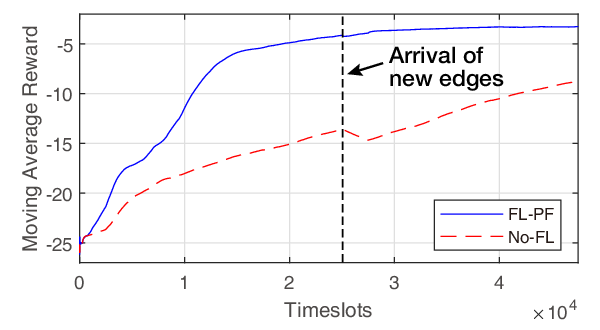}
		\label{fig:moving_avg_t3}}
	\hfil
	\caption{Moving average rewards of the edges with each task. New edges with each task having unseen scenarios arrive at ${25,}{000}$ time slots}
	\label{fig:moving_avg}
\end{figure*}

\subsection{Effectiveness to Unseen Edge Arrivals}
\label{sec:unseen_edge}
Here, we take into consideration newly arrived edges to demonstrate the efficacy of our collaborative policy learning framework when dealing with unforeseen edge arrivals.
For each task, we simulate the arrival of four edges after ${25,}{000}$ time slots; two edges are associated with scenario D and two with scenario E, as defined in Table \ref{table:task_scenarios} in Appendix \ref{appendix:experiment}.
It is important to note that these scenarios are novel to the IoT network, and as such, the policies for each task have no prior experience with them.

Fig. \ref{fig:moving_avg} illustrates the moving average rewards of FL-PF and No-FL for each task, employing a ${2,}{500}$-time slot average window for the moving average operation.
It should be noted that FL-PF is chosen as the representative algorithm among the collaborative policy learning algorithms for this comparison.
As seen in Figs. \ref{fig:moving_avg_t1}, \ref{fig:moving_avg_t2}, and \ref{fig:moving_avg_t3}, FL-PF does not experience reward degradation due to the arrival of new edges, in contrast to No-FL.
Specifically, our collaborative policy learning framework can immediately utilize the task policy located at the cloud server when a new edge with a task arrives.
On the other hand, No-FL necessitates learning a new policy for the newly arrived edge, leading to performance degradation during the initial learning phase.
These results distinctly underscore the effectiveness of our collaborative policy learning framework in managing dynamic edge arrivals in realistic IoT networks.

\begin{figure}[!t]
	\centering
	\subfloat[Learning speed.]{\includegraphics[width=0.48\linewidth]{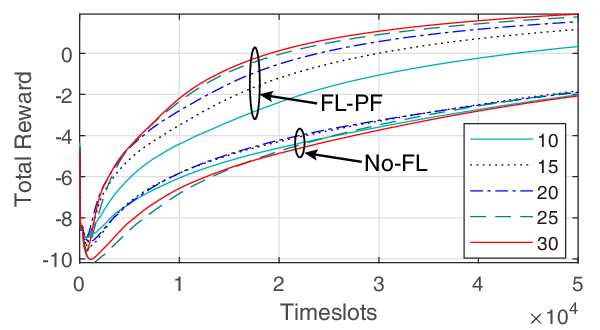}
		\label{fig:number_edges_speed}}
	\hfil
	\subfloat[Total average reward.]{\includegraphics[width=0.48\linewidth]{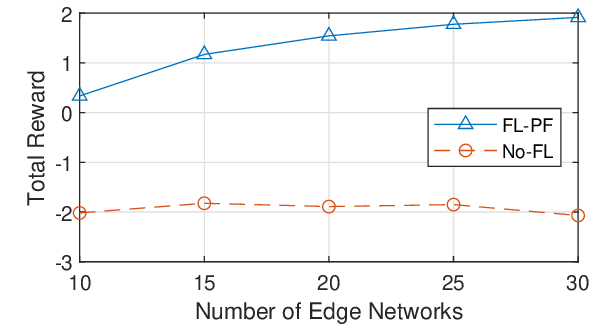}
		\label{fig:number_edges_reward}}
	\caption{Learning speed and total average rewards of FL-PF and No-FL varying the number of edges.\label{fig:impact_numbers}}
\end{figure}

\subsection{Impact of Number of Edges}
\label{sec:number_edges}

We explore the impact of the number of edges on our collaborative policy learning framework.
To this end, we show the total average rewards of FL-PF and No-FL with varying numbers of edges for each task in Fig. \ref{fig:impact_numbers}.
We adjust the number of edges from $10$ to $30$.
As federated learning involving a larger number of edges necessitates more resources, we proportionally set the maximum network bandwidth, memory, and computing resources relative to the basic setting.

In Fig. \ref{fig:number_edges_speed}, we contrast the learning speeds of FL-PF and No-FL as a function of the number of edges.
It is clear from the figure that FL-PF learns significantly faster than No-FL, which attests to the efficacy of collaborative policy learning.
Moreover, FL-PF's learning speed increases as the number of edges grows.
This suggests that our proposed framework can more rapidly learn an edge-agnostic policy when there are more edges, by capitalizing on their collective experiences through collaborative policy learning.
Conversely, no discernable trend exists in No-FL's learning speed relative to the number of edges, as each edge in No-FL must rely solely on its own experience to learn a policy.

Fig. \ref{fig:number_edges_reward} compares the total average rewards of FL-PF and No-FL at the conclusion of the simulation.
The total average reward of FL-PF increases as the number of edges grows, suggesting that a larger number of edges is beneficial for achieving greater rewards, due to the quicker learning speed.
However, no discernable trend is observed in the total average reward of No-FL in relation to the number of edges, given the absence of collaborative policy learning.
These results clearly demonstrate that our proposed collaborative policy learning framework can effectively leverage experiences from a larger number of edges.

\section{Conclusion}
\label{sec:conclusion}
In this paper, we proposed a collaborative policy learning framework for the dynamic scheduling tasks in IoT networks using FRL.
This framework effectively utilizes limited cloud resources while ensuring fair local policy aggregation across tasks.
To achieve this, we developed a task selection algorithm that maximizes the average number of participants (i.e., participating edges) in collaborative policy learning while satisfying the minimum number of edges required for each task.
We also investigated the convergence of collaborative policy learning based on this task selection algorithm.

A key enabler of the proposed framework is the edge-agnostic policy structure that we proposed for dynamic scheduling tasks, which is applicable to collaborative learning.
Our experimental results demonstrate that the proposed framework offers significant performance improvements compared to algorithms without collaborative policy learning.
Notably, the collaborative policy learning approach, when combined with our proposed task selection algorithm, achieves the best performance.
Furthermore, our results clearly illustrate the framework's adaptability to newly arrived edges and its ability to accelerate the learning speed of the policy.

\appendices
\section{Proof of Theorem 2}
\label{appendix:proof}
Since we can show the convergence rate of the DNN for task $l$ similarly to the theoretical results in \cite{ruan2021towards}.
First we derive the following theorem from Theorem 3.1 in \cite{ruan2021towards}:

\begin{theorem}
By choosing the learning rate, $\eta_r$, as $\eta_r = \frac{16E}{\xi\mathbb{E}[\sum_{n\in\calN(l)}c_n^r]}\frac{1}{rE+\gamma}$, we can obtain
\begin{equation}
	\mathbb{E}\Vert \btheta_l^r-\btheta_l^*\Vert^2\leq\frac{G}{rE+\gamma},
\end{equation}
where $E$ is the number of local epochs, 
\begin{align}
	\gamma &= \max\left\{ \frac{32E(1+N_l)L}{\xi\mathbb{E}[\sum_{n\in\calN(l)c_n^rE}]},\frac{4E^2N_l}{\mathbb{E}[\sum_{n\in\calN(l)c_n^rE}]} \right\},\\
	G &= \max\left\{ \gamma^2\mathbb{E}\Vert \btheta_l^0-\btheta_l^*\Vert^2,\left( \frac{16E}{\xi\mathbb{E}[\sum_{n\in\calN(l)c_n^rE}]} \right)^2\frac{\mathbb{E}[B_r]}{E} \right\}, \\
	B_r &= 2(2+N_l)L\sum_{n\in\calN(l)}c_n^rE\Gamma_l^n+2EV^2\sum_{n\in\calN(l)}\frac{(c_n^r)^2}{c_n}E \nonumber \\
	& \quad +\left(\frac{4(1+N_l)L+\xi}{2(1+N_l)L}\right)E(E-1)V^2\left(\sum_{n\in\calN(l)}(2+N_l)c_n^r-2N_l\right) \nonumber \\
	&\quad+\sum_{n\in\calN(l)}(c_n^r)^2E\sigma_n^2.
\end{align}
\end{theorem}

From the assumptions, we have $\mathbb{E}[B]=O(N_l^2\mathbb{E}[\frac{1}{X_l^r}|X_l^r\neq0]+\sum_{n\in\calN(l)}(c_n\sigma_n)^2+\Gamma_l)$, and $\gamma=O(N_l)$, where $X_l^r$ denotes the number of participants for task $l$ in round $r$.
Thus, $G=O(N_l^2\mathbb{E}[\frac{1}{X_l^r}|X_l^r\neq0]+\sum_{n\in\calN(l)}(c_n\sigma_n)^2+\Gamma_l)$.
Since we can easily derive these equations in similar steps to Corollary 4.0.1 in \cite{ruan2021towards}, we here omit the proofs and refer to \cite{ruan2021towards} for more details.
Given that $\mathbb{E}[\frac{1}{X_l^r}|X_l^r\neq0]\leq 1$, we can derive the theorem.

\section{Detailed Description and Scenarios of Each Task in Experiments}
\label{appendix:experiment}

In this appendix, we provide a detailed description of each task used in the experimental result section.
We also provide three different scenarios of each task in Table \ref{table:task_scenarios}.

\setlength\tabcolsep{3pt}
\setlength{\textfloatsep}{5pt}
\begin{table*}[!t]
\footnotesize
\centering
\caption{Different scenarios for each task}
\label{table:task_scenarios}
\begin{tabular}{c|c|c|c|c|c|c}
	\hline
	                              \multicolumn{7}{c}{\textbf{Wireless power transfer task (Task A)}}                               \\ \hline\hline
	          \multicolumn{2}{c|}{Scenarios}            &  \textbf{A}  &  \textbf{B}  &  \textbf{C}  &  \textbf{D}  &  \textbf{E}  \\ \hline\hline
	       \multicolumn{2}{c|}{No. IoT devices}         &      7       &      8       &      9       &      8       &      8       \\ \hline
	    \multicolumn{2}{c|}{Initial battery level}      & 20\,mJ for all & 30\,mJ for all & 40\,mJ for all & 30\,mJ for all & 40\,mJ for all \\ \hline
	        \multicolumn{2}{c|}{Charging rate}          &                     \multicolumn{5}{c}{5 mW for all}                     \\ \hline
	        \multicolumn{2}{c|}{Discharging rate}         &                \multicolumn{5}{c}{1 mW for all}                \\ \hline\hline
	                                  \multicolumn{7}{c}{\textbf{Data gathering task (Task B)}}                                    \\ \hline
	          \multicolumn{2}{c|}{Scenarios}            &  \textbf{A}  &  \textbf{B}  &  \textbf{C}  &  \textbf{D}  &  \textbf{E}  \\ \hline
	       \multicolumn{2}{c|}{No. IoT devices}         &      4       &      7       &      10      &      6      &      9      \\ \hline
	\multirow{3}{*}{\makecell{Average transmission capacity\\(samples)}} & 30 &    1 dev     &    3 devs    &    3 devs    &    2 devs    &    3 devs    \\ \cline{2-7}
	                                               & 50 &    2 devs    &    2 devs    &    4 devs    &    2 devs    &    3 devs    \\ \cline{2-7}
	                                               & 70 &    1 dev     &    2 devs    &    3 devs    &    2 devs    &    3 devs    \\ \hline
	         \multicolumn{2}{c|}{Arrival rate (samples/time slot)}          &  15 for all  &  10 for all  &  5 for all   &  10 for all   &  5 for all   \\ \hline
	        \multicolumn{2}{c|}{Maximum buffer}         &                \multicolumn{5}{c}{90 samples for all IoT devices}                \\ \hline\hline
	                             \multicolumn{7}{c}{\textbf{Radio resource scheduling task (Task C)}}                              \\ \hline
	          \multicolumn{2}{c|}{Scenarios}            &  \textbf{A}  &  \textbf{B}  &  \textbf{C}  &  \textbf{D}  &  \textbf{E}  \\ \hline
	       \multicolumn{2}{c|}{No. IoT devices}         &      4       &      9       &      20      &      6      &      12      \\ \hline
	    \multirow{3}{*}{Distance from AP (m)}      & 20 &    1 dev     &    3 devs    &    5 devs    &    2 devs    &    4 devs    \\ \cline{2-7}
	                                               & 50 &    2 devs    &    3 devs    &   10 devs    &    2 devs    &    4 devs    \\ \cline{2-7}
	                                               & 80 &    1 dev     &    3 devs    &    5 devs    &    2 devs    &    4 devs    \\ \hline
	        \multicolumn{2}{c|}{Data rate requirement (Mbps)}         &  1 for all   & 0.5 for all  & 0.2 for all  &  0.4 for all  &  0.3 for all  \\ \hline
	          \multicolumn{2}{c|}{Log-normal shadowing}            &                    \multicolumn{5}{c}{10 dB for all}                     \\ \hline
\end{tabular}
\end{table*}

\textbf{Task A: Wireless power transfer task} \textendash{}
This task aims to minimize the power outages of IoT devices cased by low battery levels \cite{lee2019contextual}.
For simple presentation, we assume, without loss of generality, that each time slot lasts for one second, and an AP wirelessly transfer power to an IoT device in each time slot.
The charging rate of each IoT device through wireless power transfer in time slot $t$, denoted by $P_m^{ch,t}$, depends on its wireless channel condition at that time slot, which typically varies with time.
If an IoT device is in an active state, its battery is discharged at a given rate.
We denote the active state of IoT device $m$ in time slot $t$ by $x_m^t$, where $1$ represents active and $0$ represents inactive.
The active state probabilistically changes based on a Markov model.
The state transition probability of IoT device $m$ from active to inactive is denoted by $p_m^{ai}$, and that from inactive to active is denoted by $p_m^{ia}$, both of which are set to $0.5$ in this task.
The battery level of IoT device $m$ in time slot $t$ is denoted by $b_m^t$.
Then, the battery level of IoT device $m$ is updated according to its active state and wireless power transfer from the AP as follows:
\begin{equation}
	b_m^{t+1}=\min\left[\max[0,b_m^t-x_m^tP_m^{dch}+q_m^tP_m^{ch,t}],B_m\right],
\end{equation}
where $P_m^{dch}$ is the discharging rate of IoT device $m$ in the active state, $q_m^t$ is the scheduling indicator of IoT device $m$ in time slot $t$, and $B_m$ is the maximum battery level.
In this task, the battery level $b_m^t$, active state $x_m^t$, and charging rate $P_m^{ch,t}$ of each IoT device in each time slot are used as state information.
The reward in each time slot is defined by the number of IoT devices whose battery levels are low (e.g., under $10\,\%$ of the maximum battery level) and that experience an outage as follows:
\begin{equation}
	r^t=-\sum_{m\in\calM}[\mathbf{1}\{b_m^t \leq B_m^{low}\} + C_m\mathbf{1}\{b_m^t=0\}],
\end{equation}
where $B_m^{low}$ is the threshold for the low battery state of IoT device $m$, and $C_m$ is the cost parameter for the outage of IoT device $m$.

\textbf{Task B: Data Gathering Task} \textendash{}
This task aims to maximize the number of gathered data samples while minimizing the dropped data samples in an IoT network \cite{kaur2020data}.
In each time slot, an IoT device is selected to transmit its data samples to the AP
We denote, for each IoT device $m$ in time slot $t$, the buffer size by $b_m^t$, the transmission capacity by $c_m^t$, and the number of arrived data samples by $d_m^t$.
The transmission capacity of IoT device $m$ in each time slot is determined by applying a floor function to a number sampled from a Gaussian random variable with mean $C_m$ and variance $9$, and the number of arrived data samples at IoT device $m$ in each time slot is sampled from a Poisson distribution with a mean of $D_m$, where $D_m$ is the arrival rate of IoT device $m$.
Then, the buffer size of IoT device $m$ is updated using $b_m^{t+1}=\min\left[\max[0,b_m^t+d_m^t-q_m^tc_m^t],B_m\right]$, where $q_m^t$ is the scheduling indicator of IoT device $m$ in time slot $t$, and $B_m$ is the maximum buffer size of IoT device $m$.
The remaining buffer size of IoT device $m$ in time slot $t$ is defined as $\bar{b}_m^t=B_m-b_m^t$.
If the buffer overflows, the exceeded data samples are dropped, and the number of dropped data samples of IoT device $m$ in time slot $t$ is denoted by $e_m^t$.
The state information for the problem includes the remaining buffer size, $\bar{b}_m^t$, and transmission capacity, $c_m^t$, of each IoT device in each time slot.
The reward in each time slot is defined as the difference between the number of gathered data samples and the number of dropped data samples in the time slot, which is computed as:
\begin{equation}
	r^t=\sum_{m\in\calM}q_m^t\min[c_m^t,b_m^t+d_m^t]-e_m^t.
\end{equation}

\textbf{Task C: Radio Resource Scheduling Task} \textendash{}
This task aims to minimize the transmission power at an AP while satisfying the minimum average data rate requirements of IoT devices \cite{lee2019resource}.
Hence, we refer the readers to \cite{lee2019resource} for more details.

\bibliographystyle{IEEEtran}
\bibliography{IEEEabrv,mybib}

\end{document}